\ifx\pdfoutput\undefined\else\pdfoutput=1\fi
\documentclass[11pt]{article}

\usepackage[utf8]{inputenc}
\usepackage[T1]{fontenc}
\usepackage[margin=1in]{geometry}
\usepackage{microtype}
\usepackage{graphicx}
\usepackage[table,xcdraw]{xcolor}
\usepackage{rotating}
\usepackage[font=small,labelfont=bf]{caption}
\usepackage{subfig}
\usepackage{amsmath,amssymb}
\usepackage{booktabs}
\usepackage{array,multirow}
\usepackage{colortbl}
\usepackage{siunitx}
\usepackage[ruled,vlined]{algorithm2e}
\usepackage[numbers,sort&compress]{natbib}
\usepackage[hidelinks]{hyperref}

\title{Structured-Condensed Prompt Tuning in Vision-Language Models for Fine-grained Image Recognition}

\author{%
Xinda Liu\textsuperscript{1} \quad
Qinyu Zhang\textsuperscript{1} \quad
Weiqing Min\textsuperscript{2}\\
Guohua Geng\textsuperscript{1} \quad
Shuqiang Jiang\textsuperscript{2}\\[0.5em]
\small \textsuperscript{1}School of Information Science and Technology, Northwest University, Xi'an, Shaanxi, China\\
\small \textsuperscript{2}Laboratory of Intelligent Information Processing, Institute of Computing Technology, Beijing, China\\
\small \texttt{liuxinda@nwu.edu.cn, zhangqinyu@stu.nwu.edu.cn, minweiqing@ict.ac.cn}\\
\small \texttt{ghgeng@nwu.edu.cn, sqjiang@ict.ac.cn}
}

\date{}

\begin{document}

\maketitle

\begin{abstract}
Fine-grained image recognition poses a significant challenge due to the substantial expertise and effort required for manual annotation. Vision-language models (VLMs) like CLIP provide a compelling zero-shot alternative, reducing reliance on extensive labeled data. However, their ability to capture subtle distinctions remains limited, leading to subpar recognition performance. While prompt tuning has proven effective for adapting VLMs, most existing methods treat class labels as isolated, discrete entities, overlooking the rich semantic relationships between them. This oversimplified assumption limits the model's ability to capture hierarchical dependencies and inter-class correlations---both critical for distinguishing visually similar categories. The problem is especially acute in fine-grained classification, where accurate recognition depends on understanding complex label semantics. To address this, we propose Structured-Condensed Prompt Tuning (SCPT), which enhances semantic structure modeling in prompt learning. Specifically, we introduce Semantic Relation Encoding (SRE) to explicitly model inter-class semantic topology and encode structured label relationships. In parallel, we design a Semantic Condensation loss (ScLoss) to suppress redundant supervision and extract discriminative components from the global semantic space. Together, these components significantly improve semantic alignment and fine-grained discrimination. Extensive experiments on 14 fine-grained benchmarks show that SCPT effectively mitigates semantic ambiguity and achieves state-of-the-art performance in both few-shot and base-to-novel generalization settings.
\end{abstract}

\noindent\textbf{Keywords:} Prompt tuning; Vision-language models; Semantic relation; Few-shot learning
\section{Introduction}
\begin{figure}[ht]
    \centering
    \includegraphics[width=0.65\linewidth]{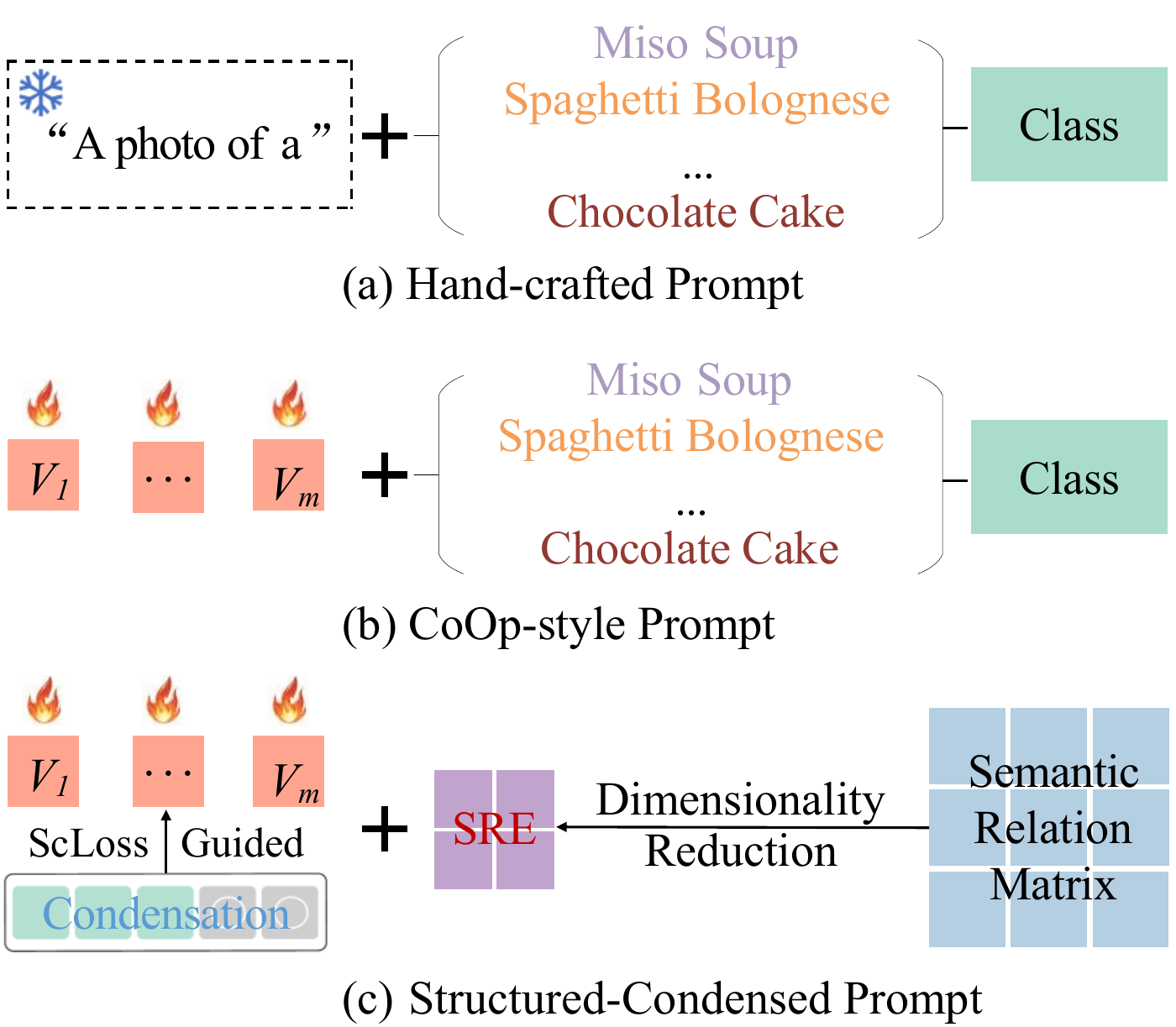}
    \caption{From static to semantically structured prompts: a paradigm shift in prompting mechanisms.}
    \label{fig-mo}
\end{figure}

Fine-grained image recognition  (FGIR) plays a pivotal role in scenarios demanding precise differentiation between visually similar subcategories,  rich image captioning \cite{LIU2024110452,LI2025111806,chang2023changes},  image generation \cite{SUN2024109962,xu2023txt2img},  food recognition \cite{XIAO2024112134,liu2024convolution, zhou2024synthesizing},  and food recommendation \cite{QIAO2025104877,min2023large}. The essential challenge stems from the requirement for expert-level annotation precision,  which necessitates a nuanced understanding of subtle visual differences,  often demanding domain-specific knowledge \cite{vu2023instance}. The exorbitant cost associated with acquiring such human-expert annotations has emerged as a critical bottleneck,  that fundamentally constrains the development of FGIR systems in novel domains.

Vision-language models  (VLMs),  such as CLIP,  offer a promising solution by leveraging large-scale web data for zero-shot learning,  thereby mitigating the reliance on extensive manual annotations \cite{radford2021learning}. CLIP employs contrastive learning to align text and image representations,  enabling category classification without task-specific training. 
While VLMs demonstrate strong generalization in broad categories,  this advantage diminishes notably in fine-grained recognition contexts,  predominantly stemming from the model's constrained capacity to discern nuanced inter-class variations.

Prompt tuning adapts task-specific prompts to better leverage the latent knowledge in vision-language models (VLMs), driving strong performance on downstream tasks. It falls into two categories: visual and textual prompt tuning, with the latter optimizing text inputs for improved alignment with visual representations.  In this work,  we focus exclusively on textual prompt tuning,  omitting modifications to the visual encoder. Instead of relying on manually crafted prompts,  textual prompt tuning replaces static templates with learnable context vectors \cite{zhou2022learning, zhou2022conditional},  which are dynamically optimized using a small set of training samples to enhance image-text alignment,  thereby alleviating data scarcity constraints.

However,  existing prompt tuning approaches treat category labels as independent,  discrete entities,  failing to leverage the rich semantic relationships among them. This oversimplification limits the model's capacity to capture hierarchical dependencies and inter-class correlations,  which are crucial for distinguishing visually similar categories. The challenge is particularly pronounced in fine-grained classification,  where effective recognition hinges on understanding the intricate semantic structure underlying label relationships.

We address this limitation by introducing \textbf{Structured-Condensed Pro\-mpt Tuning} (SCPT), which formulates a structure-aware semantic alignment mechanism within the CoOp-style prompt tuning framework. Instead of relying on discrete class tokens, SCPT embeds Semantic Relation Encoding (SRE) into the prompt space, preserving the global semantic topology among labels through pairwise relational distances, as shown in Figure  \ref{fig-mo} (c). This enables vision-language models to reason over structured category representations, where each category embedding explicitly encodes its semantic relations to other categories in a shared embedding space, rather than memorizing disconnected labels. We further propose Semantic Condensation loss (ScLoss) to enhance semantic focus by suppressing intra-class noise and amplifying discriminative cues essential for classification.  Together, SRE and ScLoss form a unified framework that captures both inter-class structure and intra-class focus, enhancing both few-shot adaptation and base-to-novel generalization. Experimental results show that SCPT consistently outperforms state-of-the-art prompt tuning methods,  achieving substantial gains in both few-shot learning and base-to-novel generalization.

In summary,  the key contributions of this paper are as follows:

(1) We propose a structure-aware prompt tuning method, Structured-Condensed Prompt Tuning (SCPT), which enhances the semantic structure modeling in vision-language models for FGIR. This approach explicitly captures the inter-class semantic relationships and improves the model's ability to distinguish subtle differences between visually similar categories.

(2) We present SRE to model the inter-class semantic topology by encoding structured label relationships. This method leverages pairwise relational distances to preserve the global semantic structure, enabling the model to better understand the hierarchical dependencies and correlations between classes.

(3) We design ScLoss to suppress redundant supervision signals and extract discriminative components from the global semantic space. This loss function enhances the model's focus on task-relevant semantics, improving both few-shot adaptation and generalization to novel classes.

(4) We conduct extensive experiments on 14 diverse FGIR benchmarks. SCPT demonstrates superior performance compared to existing prompt tuning methods. It achieves significant gains in both few-shot learning and base-to-novel generalization tasks, establishing a new state-of-the-art in this domain.

\section{Related Works}
\subsection{Vision-Language Models  (VLMs)}
VLMs such as CLIP \cite{radford2021learning},  ALIGN \cite{jia2021scaling},  and BLIP \cite{li2022blip} leverage large image-text datasets to achieve superior discriminative power and generalization over traditional single-modality models. By aligning visual and textual information through self-supervised learning,  these models excel in zero-shot tasks and transfer their learned representations to downstream applications like image retrieval \cite{baldrati2022effective, sain2023clip},  image segmentation \cite{wang2022cris, luddecke2022image},  and visual question answering \cite{eslami2023pubmedclip}. Although CLIP has shown strong zero-shot performance,  fine-tuning for specific tasks can enhance its effectiveness,  especially in resource-limited settings. Recent works \cite{zhou2022learning, zhou2022conditional} have explored fine-tuning VLMs for few-shot image recognition,  addressing challenges like data scarcity and computational constraints. In this paper,  we propose a novel text-prompt tuning method to adapt CLIP for few-shot fine-grained visual recognition,  enhancing the model's flexibility and applicability.

\subsection{Prompt Tuning for VLMs}
Prompt tuning seeks to enhance the adaptability of VLMs by introducing a limited number of learnable parameters,  thereby allowing the generic features acquired during pre-training to be better aligned with relevant downstream tasks. The fundamental concept is to integrate category information through learnable textual or visual prompts,  replacing handcrafted templates such as ``a photo of a {classname}''. These prompts are optimized via backpropagation,  while the pre-trained model remains frozen. Existing prompt tuning methodologies often improve model performance by leveraging learnable parameters across various modalities while incorporating additional semantic information. For instance,  CoOp\cite{zhou2022learning} replaces the handcrafted templates in CLIP with learnable soft prompts that integrate class names,  resulting in substantial advancements in few-shot classification tasks. CoCoOp\cite{zhou2022conditional} further enhances the generalization capability of the CoOp model by merging image features with textual soft prompts. Moreover,  methods such as MaPle \cite{khattak2023maple},  and PromptSRC\cite{khattak2023self} also introduce learnable visual prompts to bolster model performance. More recently, DGPrompt \cite{zheng2025dgprompt} proposes a dual-guidance prompt generation framework that jointly exploits textual semantic cues and visual structural information to enhance cross-modal alignment in vision-language models.  ProDa \cite{lu2022prompt} optimizes the direction of prompt updates while discarding conflicting adjustments to constrain the semantic differences between learnable text features and general text semantics. Conversely,  KgCoOp \cite{yao2023visual} limits the distance between learnable text features and their general semantic counterparts,  ensuring that specific semantics remain closely aligned with broader concepts. Building upon KgCoOp,  TCP \cite{yao2024tcp} integrates class-related general semantic knowledge extracted from CLIP into deeper layers of the text encoder,  thereby enhancing the encoder's capacity to differentiate between classes.  Related to structured semantics, graph-based methods rely on explicit graph construction and additional supervision to model inter-class relationships, whereas SCPT leverages the implicit semantic structure of CLIP's pretrained embedding space without introducing extra graphs or annotations.

Despite the advancements made by existing prompt tuning approaches in fine-grained image recognition,  many still utilize original categories as classification labels,  inadequately addressing the semantic limitations inherent in FGIR tasks.  Beyond prompt-based adaptation, recent work has explored test-time adaptation for vision-language models. Bayesian Test-Time Adaptation \cite{zhou2025bayesian} frames adaptation as Bayesian inference to handle distribution shifts at inference time, but does not explicitly model the semantic granularity or structural relationships required for fine-grained recognition.
 Therefore,  when processing fine-grained categories,  CLIP may lack sufficient associative semantics learned from its training data,  leading to suboptimal generalization performance. In contrast,  the proposed method explicitly models semantic structural relationships and enhances FGIR performance by extracting discriminative components from the global semantic space.

\section{Method}

\begin{figure*}[ht]
    \centering
    \setlength{\abovecaptionskip}{0cm} 
    \includegraphics[width=\textwidth]{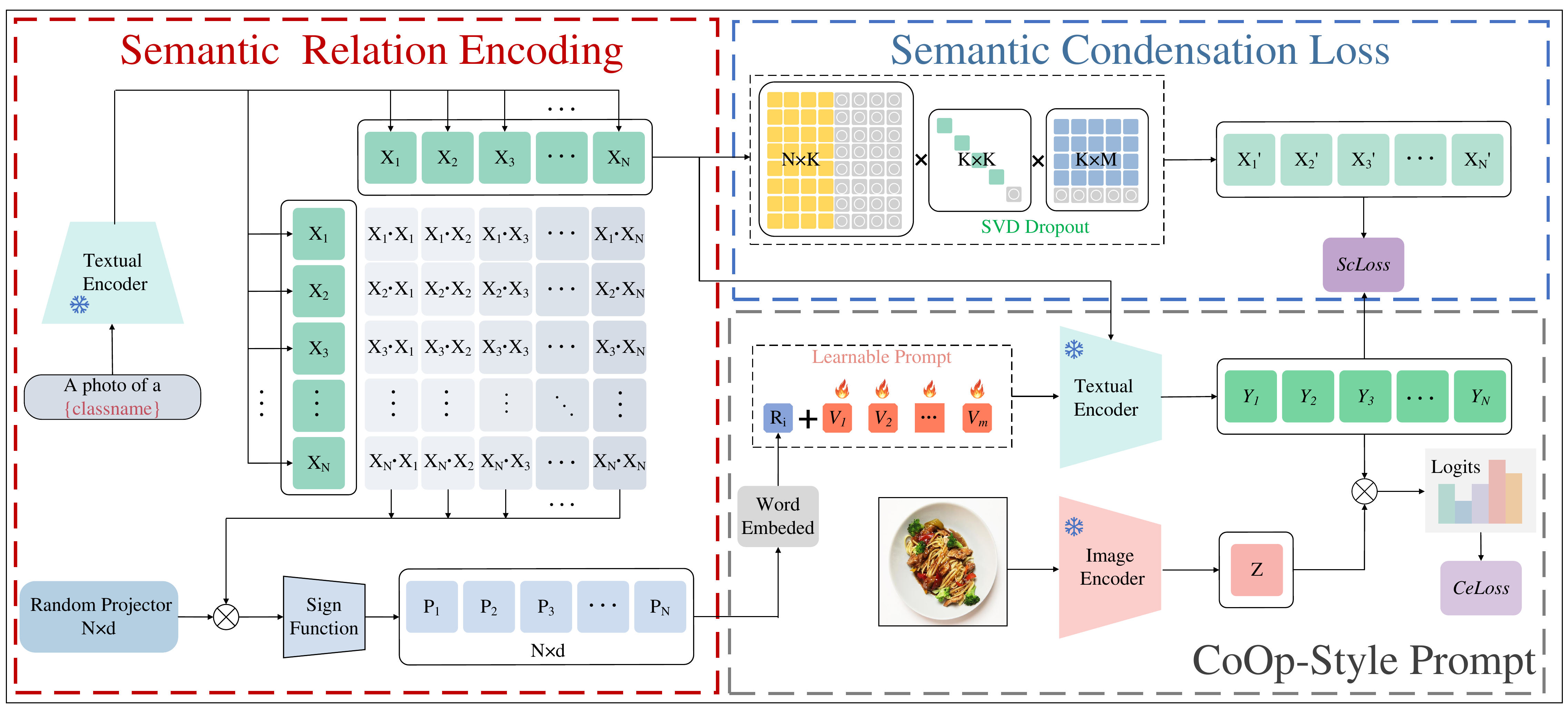}
    \caption{Overview of the proposed Structured-Condensed Prompt Tuning (SCPT) framework.}
    \label{fig-framework}
\end{figure*}
The proposed method is grounded in the observation that employing isolated,  discrete category labels within prompt templates for prompt tuning underutilizes the rich semantic information inherent in these labels. This drawback is especially pronounced in fine-grained image recognition tasks. To mitigate this issue,  we introduce Structured-Condensed Prompt Tuning  (SCPT),  which enhances the semantic expressiveness of fine-grained label prompts by incorporating Semantic Relation Encoding  (SRE) alongside a Semantic Condensation loss  (ScLoss),  as illustrated in Figure  \ref{fig-framework}. In the following sections,  we first provide a brief review of the CoOp-style prompt method,  followed by a detailed description of the proposed SCPT framework.

\subsection{CoOp-style Prompt Tuning}
Existing CoOp-style prompt tuning algorithms utilize the powerful 
CLIP as their backbone. CLIP consists of a Transformer-based text encoder  and a vision encoder based on either Vision Transformers \cite{alexey2020image} or Residual Networks \cite{he2016deep},  where the vision encoder generates visual embeddings from images,  and the text encoder transforms text prompts into textual embeddings. During training,  CLIP employs contrastive loss to align these embeddings,  facilitating effective zero-shot inference. For zero\--shot classification,  CLIP compares image features with class weights generated from descriptive text,  enabling classification without requiring specific category training. Formally,  let the image feature extracted by the image encoder be denoted as $\mathbf{Z}$,  and let $\mathbf{X} = \{ X_i \}_{i=1}^{N}$ represent the class weight vectors generated by the text encoder,  where $ N $ is the number of classes. Each class is associated with a handcrafted prompt template,  which is first transformed into a vectorized textual token using the word embedding function $ e (.) $. Specifically,  for the $ i $-th class,  the textual token is given by:  

\begin{equation}
    t_i = e (\text{``a photo of [CLASSNAME]''}).
\end{equation}
Subsequently,  the text encoder $ \theta $ maps these vectorized textual tokens into class-level embeddings: \begin{equation}
    X_i = \theta (t_i).
\end{equation}
Finally,  the prediction probability is defined as:

\begin{equation}
    p (y = i | x) = \frac{\exp\left (\frac{\mathrm{sim} (X_i,  Z)}{\tau}\right)}{\sum_{j=1}^{N} \exp\left (\frac{\mathrm{sim} (X_j,  Z)}{\tau}\right)}, 
\end{equation}
where $ \tau $ is a temperature parameter learned by CLIP and $ \mathrm{sim} (\cdot,  \cdot) $ denotes cosine similarity.

To enhance CLIP's performance on downstream tasks,  CoOp replaces the handcrafted prompt template with learnable prompts. By training on a small number of label-relevant images,  CoOp improves the discriminability of class-specific textual embeddings. Specifically,  CoOp introduces $ m $ learnable context vectors $ V = \{ v_1,  v_2,  \dots,  v_m \} $. The class label embedding $ c_i $ for the $ i $-th class is concatenated with the context vectors to form the prompt $ V^{*}_i = \{ c_i,  v_1,  v_2,  \dots ,  v_m \} $. This prompt $ V^{*}_i $ is then fed into the text encoder $ \theta $,  yielding the textual class embedding $ Y_i = \theta (V^{*}_i) $. Finally,  the textual embeddings for all classes are defined as $ \mathbf{Y} = \{ Y_i^{coop} \}_{i=1}^{N} $. 

CoOp optimizes the learnable context vectors $ V $ by minimizing the contrastive loss between the image embedding $ z $ and its corresponding class embedding $ \mathbf{Y} $:
\begin{equation}
    L_{ce} = -\sum_{\mathbf{z} \in \mathbf{Z}} \log \frac{\exp \left ( \mathrm{sim}\left ( \mathbf{Z},  \mathbf{Y}_{gt} \right) / \tau \right)}{\sum_{i=1}^{t} \exp \left ( \mathrm{sim}\left ( \mathbf{Z},  \mathbf{Y}_i \right) / \tau \right)}, 
\end{equation}
where $ gt $ is the corresponding label of the image embedding,  $ \mathbf{Z} $ is the complete set of embeddings.

To further enhance textual semantics and improve the discriminability of CoOp,  previous works \cite{yao2024tcp} have explored refining textual embeddings by reducing their dimensionality and incorporating class-aware tokens into the text encoder. Building on these approaches,  SCPT applies this enhancement within the text encoder,  facilitating the generation of more discriminative textual representations.

\subsection{Structured-Condensed Prompt Tuning} \label{sec-coop}
Unlike existing CoOp-style prompt tuning methods,  the proposed SCPT framework enhances the modeling of label semantic structure by jointly leveraging two complementary components.  SRE explicitly captures the inter-class semantic topology,  replacing isolated category labels with structure-aware embeddings that encode pairwise semantic relations among categories. ScLoss compresses supervision signals by suppressing redundancy and extracting discriminative components from the global semantic space. As illustrated in Figure \ref{fig-framework},  these two modules work in tandem to improve semantic alignment and fine-grained discriminability. We detail each component in the following sections.

\subsubsection{Semantic Relation Encoding} 
 In this work, we define structured category representations as category embeddings that preserve global inter-class semantic topology, rather than treating each category as an independent token.
The CLIP model,  with its strong zero-shot generalization capabilities,  is employed to extract semantic features for category classification. For fine-grained image recognition tasks with $ N $ categories,  we construct category-specific prompts $ c_i $ in the form ``a photo of a {$c_i$}",  which are passed through CLIP's text encoder to generate category embeddings,  denoted as $ \mathbf{X}$. Each $ X_i $ represents the semantic feature of the $ i $-th category. To quantify semantic relation between categories,  we compute a similarity matrix using cosine similarity:
\begin{equation}
    S_{ij} = \frac{\mathbf{X}_i \cdot \mathbf{X}_j}{\|\mathbf{X}_i\| \|\mathbf{X}_j\|},  \quad \forall i,  j \in \{1,  2,  \ldots,  N\}, \label{eq1}
\end{equation}
where $ \mathbf{X}_i $ and $ \mathbf{X}_j $ are the feature vectors of the $ i $-th and $ j $-th categories. Each row $ S_i $ in the similarity matrix captures the semantic similarity between the $ i $-th category and all others. This design stems from the research that CLIP's text embeddings inherently encode taxonomic relationships,  and preserving such relations through similarity constraints Eq.(\ref{eq1}) helps prevent semantic distortion during prompt tuning.

However,  due to CLIP's text prompt length limitations \cite{zhang2024long},  directly using the similarity matrix $ S $ as category labels is not feasible. Inspired by the Johnson--Lindenstrauss  (JL) Lemma \cite{johnson1984extensions},  which states that a set of $ N $ points in a high-dimensional space can be embedded into a lower-dimensional space $ \mathbb{R}^d $  ($ d \geq \mathcal{O} (\epsilon^{-2} \log N) $) while approximately preserving pairwise distances,  we propose a signed random projection method for compact semantic representation. Given a similarity matrix $ S \in \mathbb{R}^{N \times N} $,  we generate a binary embedding $ P = \{P_i\}_{i=1}^N \in \{0, 1\}^{N \times d} $ as follows:  

\begin{equation}
   P = \text{sign}\left (S W^\top\right),  \quad 
\end{equation}
where $ W \in \mathbb{R}^{d \times N} $ is a random projection matrix whose entries are drawn i.i.d. from $ \mathcal{N} (0, 1) $. Each row of $W$ defines a hyperplane that randomly partitions the semantic space,  and the sign operation encodes whether category embeddings lie on the positive side of these hyperplanes. This binarization compresses semantic relations into a Hamming space where similar categories share overlapping bit patterns. The sign function $\text{sign} (\cdot)$ applies element-wise binarization,  mapping positive values to $ 1$ and non-positive values to $ 0$. Geometrically,  each row vector $ W_i \in \mathbb{R}^{1 \times N} $ defines a randomly oriented hyperplane that partitions the semantic space,  and $ P_i \in \{0, 1\}^d $ encodes the relative positioning of categories with respect to $ W_i $. According to JL Lemma,  the number of parameter $ d $ is defined as $ d = \lceil \log_2 (N) + d_{free} \rceil $,  where $ d_{free} $ introduces additional space to enhance randomness. This ensures the vectors remain independent and non-repetitive after dimensionality reduction. 

Similar to the CoOp approach,  we first input the binary encoding $P$ into the word embedding function $e (\cdot)$ to generate the SRE,  denoted as $R = \{R_i^N\}$,  where $R_i=e (P_i)$. Subsequently,  we combine the SRE with a learnable context vector to construct a semantic distance-aware prompt vector:
\begin{equation}
    V_i = \{R_i,  v_1,  v_2,  \dots,  v_m\}.
\end{equation}
This approach efficiently encodes relational semantics while preserving inter-class structure in the lower-dimensional space.

\subsubsection{Semantic Condensation loss}

KgCoOp \cite{yao2023visual} demonstrates that aligning learnable prompts with handcrafted prompt-generated embeddings via an MSE regularizer helps extract discriminative components from the global semantic space,  thereby improving classification performance. However,  this approach may simultaneously introduce redundant or noisy signals---particularly problematic in fine-grained settings,  where class semantics are narrowly scoped and visually diverse. Such coarse supervision can mislead the optimization process and hinder the learning of compact,  class-specific representations. Motivated by this hypothesis,  we propose a straightforward yet effective strategy: Specifically,  let $\mathbf{X}\in R^{N\times M}$ denote the text embeddings generated from handcrafted prompts and $\mathbf{Y}\in R^{N\times M}$ represent the embeddings obtained from learnable prompts. Treating $ \mathbf{X} $ as a semantic data matrix and applying singular value decomposition  (SVD) to remove noise. By preserving dominant singular values and eliminating smaller ones,  we aim to denoise the embeddings. Specifically,  we perform SVD on $ \mathbf{X} $:  

\begin{equation}  
  \mathbf{X} = \mathbf{U} \mathbf{\Sigma} \mathbf{V}^\top.  
\end{equation}  
Here,  $ \mathbf{\Sigma} =  (\sigma_1,  \sigma_2,  ...,  \sigma_r) $ represents the singular values arranged in descending order,  i.e.,  $ \sigma_1 \geq \sigma_2 \geq ... \geq \sigma_r \geq 0 $,  where $ r $ denotes the rank of the matrix. We retain only the top $ K $ singular values,  yielding:  

\begin{equation}  
    \left\{  
\begin{aligned}  
    \mathbf{U}_{K} &= \mathbf{U}_{[:,  :K]} \in \mathbb{R}^{N \times K},  \\  
    \mathbf{\Sigma}_{K} &= \text{diag} (\mathbf{\Sigma}_{[:K]}) \in \mathbb{R}^{K \times K},  \\  
    \mathbf{V}_{K}^\top &= \mathbf{V}^\top_{[:K,  :]} \in \mathbb{R}^{K \times M}.  
\end{aligned}  
\right.  
\end{equation}  
The denoised embedding matrix $ \mathbf{X}^{'} $ is then reconstructed as:  

\begin{equation}  
      \mathbf{X}^{'} = \mathbf{U}_{K} \mathbf{\Sigma}_{K} \mathbf{V}_{K}^\top.  
\end{equation}  
Our goal is to mitigate the model's tendency to forget general semantics by minimizing the divergence between inter-class and generic semantic knowledge:
\begin{equation}
    L_{\text{sc}} = \frac{1}{N} \sum_{i=1}^{N} \| Y_i - X_i^{'} \|_2^2, 
\end{equation}
where $ \| \cdot \|_2 $ is the Euclidean norm,  and $ N_c $ is the number of seen classes. Finally,  we integrate the standard contrastive loss $ L_{\text{ce}} $ with our Semantic Condensation loss $ L_{\text{sc}} $ to define the overall objective function:
\begin{equation}
   L = L_{\text{ce}} + \lambda L_{\text{sc}},  
\end{equation}
where $ \lambda $ controls the contribution of $ L_{\text{sc}} $ within the total loss.

As illustrated in Figure  \ref{fig-svdmotavation} (a),  applying this denoising process improves the MSE performance compared to using the original embeddings. However,  the results also indicate that the choice of $ K $ significantly influences the outcome across different datasets. This suggests that SVD-based denoising is highly sensitive to the selection of $ K $,  raising an important question: \textbf{Can a robust mechanism be devised to determine the optimal value of $ K $?}

\subsubsection{Optimal Value of \texorpdfstring{$K$}{K}}
\begin{figure*}[ht]
    \centering
    \setlength{\abovecaptionskip}{0cm} 
    \includegraphics[width=\textwidth]{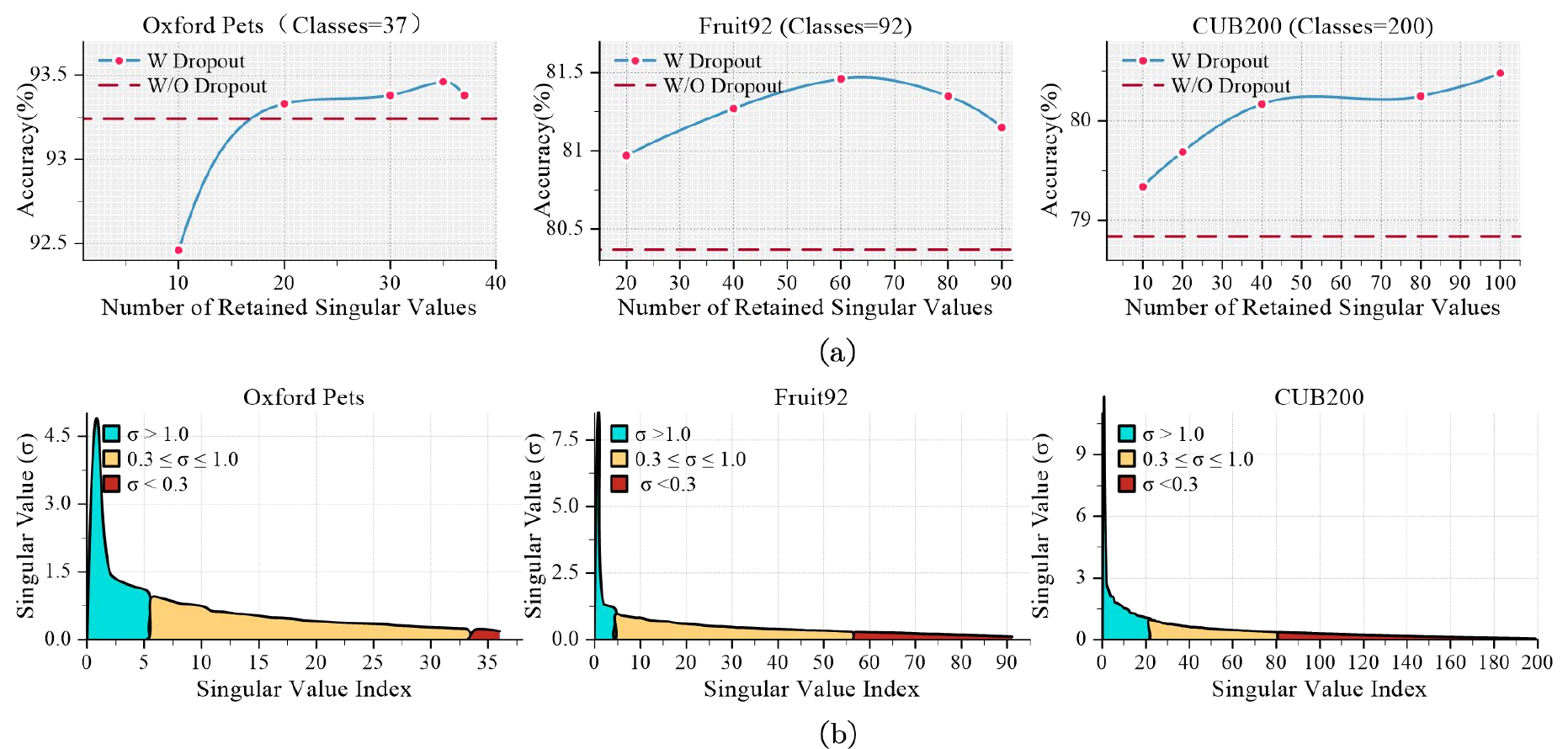}

    \caption{Comparative Analysis of SVD Dropout Efficacy: Accuracy Dynamics and Singular Value Distribution. (a)Accuracy variation across three datasets under varying principal component retention levels  $K$. Dashed red lines denote baseline performance without SVD Dropout. (b)Post-decomposition singular value magnitude distribution of semantic embeddings.}
    \label{fig-svdmotavation}
\end{figure*}
To explore this question,  we visualize the distribution of nonzero singular values after performing SVD on semantic embeddings from various fine-grained datasets,  as shown in Figure  \ref{fig-svdmotavation}. Compared to Figure  \ref{fig-svdmotavation} (a),  a consistent pattern emerges: despite differences in matrix rank,  the optimal number of retained singular values tends to fall between the mid-range and lower-range singular values. This suggests that smaller singular values predominantly capture noise,  with the smallest ones being particularly susceptible to noise-induced distortions. 

To further substantiate this observation,  We assume that $ \mathbf{X} \in \mathbb{R}^{N \times M} $ is a purely noisy Gaussian random matrix,  where each element is independently and identically distributed  (i.i.d.) with zero mean and variance $ \sigma^2 $.  The covariance matrix of $\mathbf{X}$ can be expressed as:
\begin{equation}
    \mathbf{S}=\frac{1}{M}XX^T.
\end{equation}
According to the Marchenko-Pastur theorem \cite{yaskov2016short},  when $N, M\rightarrow\infty$,  and $\frac{M}{N}\rightarrow r\in (0, 1]$. The singular values of the sample covariance matrix follow a characteristic distribution given by:  
\begin{equation}
    p (\sigma) = \frac{1}{2\pi r\sigma^2}  (\sigma_{\max} - \sigma) (\sigma - \sigma_{\min}),  \quad \sigma_{\min} \leq \sigma \leq \sigma_{\max}, 
\end{equation}  
where $ \sigma_{\min} $ and $ \sigma_{\max} $ denote the lower and upper bounds of the singular value spectrum,  determined by the data dimensions $ N,  M $ and the noise variance $ \sigma^2 $. Notably,  the density of singular values near $ \sigma_{\min} $ is markedly high,  indicating that most singular values in a pure noise matrix reside in this region. Conversely,  singular values exceeding $ \sigma_{\max} $ are statistically improbable under a pure noise assumption,  implying that they are more likely to correspond to meaningful semantic structures rather than random fluctuations.  

Building on this theoretical analysis,  we propose an adaptive filtering mechanism that removes singular values below a predefined threshold,  effectively reducing noise while preserving as much semantic information as possible. Notably,  the optimal value of $ K $ tends to shift slightly to the right of the theoretical optimum. This shift occurs because some smaller singular values may still carry semantic information. To address this,  we design a probabilistic selection function $ P (\sigma_i) $ as follows:  

\begin{equation}  
P (\sigma_i) =  
\begin{cases}  
1,  & \text{if } \sigma_i \geq \tau,  \\  
e^{  (\sigma_i - \tau)^2-N_{\text{below}}},  & \text{otherwise}.  
\end{cases} ,  
\end{equation}  
where $ \tau $ represents the threshold,  and $ N_{\text{below}} $ denotes the number of singular values below $ \tau $. This formulation ensures that singular values exceeding the threshold are always retained,  while those closer to the threshold have a higher probability of being preserved. The term $ e^{-N_{\text{below}}} $ serves as a regularization factor,  reflecting the observation that a larger number of small singular values increases the likelihood of noise. Consequently,  when more small singular values are present,  the probability of selecting any individual one is reduced,  reinforcing the suppression of noise-dominated components. Thus,  the final selection of $ K $ is determined as the index of the $ i $-th singular value that is retained based on the probabilistic selection function,  formulated as:  

\begin{equation}  
K = \arg\max_{i} \{ P (\sigma_i) \neq 0 \}.  
\end{equation}  
This ensures that $ K $ corresponds to the largest index among the selected singular values,  striking a balance between noise reduction and semantic preservation. 

\begin{algorithm}[h]
\caption{Training of Structured-Condensed Prompt Tuning (SCPT)}
\label{alg:scpt}
\SetKwInOut{Input}{Input}
\SetKwInOut{Output}{Output}

\Input{Training data $\{(x,y)\}$, class names $\{c_i\}_{i=1}^N$, frozen encoders $f_{\mathrm{img}}, f_{\mathrm{txt}}$, projection matrix $W$, hyperparameters $\lambda, \tau$}
\Output{Optimized context vectors $V$}

\tcp{Initialization}
Encode handcrafted prompts to obtain text features $X$\;
Compute class similarity matrix $S$ from $X$\;
Generate Semantic Relation Encoding (SRE):
$P = \mathrm{sign}(S W^\top), \quad R_i = e(P_i)$\;

\tcp{Optimization}
\While{until convergence}{
    Construct structured prompts $V_i = \{R_i, V\}$ and encode $Y_i = f_{\mathrm{txt}}(V_i)$\;
    Extract image features $z = f_{\mathrm{img}}(x)$\;
    Compute logits via cosine similarity\;
    Compute contrastive loss $\mathcal{L}_{\mathrm{ce}}$\;
    Compute Semantic Condensation loss $\mathcal{L}_{\mathrm{sc}}$ via SVD-based filtering with threshold $\tau$\;
    Update $V$ by minimizing $\mathcal{L}_{\mathrm{ce}} + \lambda \mathcal{L}_{\mathrm{sc}}$\;
}
\end{algorithm}

SCPT introduces a structured prompt tuning strategy that jointly leverages inter-class relational encoding and intra-class semantic compression. SRE encodes inter-class relations via randomized hyperplane projections that preserve CLIP's semantic topology,  while ScLoss distills task-critical semantics through singular value truncation. Algorithm~\ref{alg:scpt} provides a unified view of how SRE and ScLoss are integrated into the SCPT framework. Their coordinated operation, with SRE expanding inter-class discriminability and ScLoss contracting intra-class redundancy, establishes a robust foundation for fine-grained adaptation.

\section{Experiments}
To evaluate the effectiveness of the proposed method,  
 we follow standard practice on a series of downstream tasks,  including few-shot experiments,  base-to-novel generalization,  and ablation study.

\subsection{Experimental Setup}
\noindent\textbf{Datasets:}
We employed 14 fine-grained classification datasets including Dog Breed  \cite{zou2020new},  Oxford Pets \cite{parkhi2012cats},  Oxford Flowers \cite{nilsback2008automated},  Stanford Cars \cite{krause20133d},  Web Cars \cite{sun2021webly},  Stanford Dogs \cite{KhoslaYaoJayadevaprakashFeiFei_FGVC2011, imagenet_cvpr09} ,  Fruit92 \cite{hou2017vegfru} ,  CUB200 \cite{wah2011caltech}, Food101 \cite{bossard2014food},  Food172 \cite{VireoFood172},  Food200 \cite{min2019ingredient},  Food500\cite{min2020isia},  FGVC Aircraft \cite{maji2013fine} and Veg200 \cite{hou2017vegfru}.

\noindent\textbf{Implementation Details:}
For all experiments,  we utilize ViT-B/16 as the vision backbone. In the SCPT model,  the context length for learnable prompts is fixed at 4, and the context vectors are initialized from a Gaussian distribution ($\sigma=0.02$),  threshold $\tau$ is 0.3. The hyperparameter $\lambda$ is 8.0,  same as KgCoOp. The models are trained on the RTX 4090 GPU with 24GB memory using the SGD optimizer. The training process spans 50 epochs with a batch size of 32. The initial learning rate is set to 2e-3 and dynamically adjusted according to the cosine annealing schedule. All comparisons are conducted under a consistent experimental protocol, with method-specific settings applied only when required by the original implementation.

\noindent\textbf{SOTA Methods:}
We compared SCPT with the zero-shot inference capability of CLIP \cite{radford2021learning} (ICML 2021) and conducted a comprehensive comparison with recent textual prompt tuning methods. e.g.,  CoOp \cite{zhou2022learning} (IJCV 2022),  CoCoOp \cite{zhou2022conditional} (CVPR 2022),  KgCoOp \cite{yao2023visual} (CVPR 2023),  TCP \cite{yao2024tcp} (CVPR 2024),  ProText \cite{khattak2025learning} (AAAI 2025) and ATPrompt \cite{li2024advancing} (ICCV 2025).

\begin{table}[!htbp]
\centering
\footnotesize
\setlength{\tabcolsep}{0.85pt}
\caption{Comparing SCPT to other prompt tuning methods in few-shot learning (\%).}
 
\label{fewshot}
\renewcommand{\arraystretch}{1}

\begin{tabular}{lccccccccccccccc}\toprule

Method &\begin{tabular}[c]{@{}c@{}}\rotatebox{70}{Dog Breed}\end{tabular} & \begin{tabular}[c]{@{}c@{}}\rotatebox{70}{Oxford Pets}\end{tabular} & \begin{tabular}[c]{@{}c@{}}\rotatebox{70}{Oxford Flowers}\end{tabular} & \begin{tabular}[c]{@{}c@{}}\rotatebox{70}{Stanford Cars}\end{tabular} & \begin{tabular}[c]{@{}c@{}}\rotatebox{70}{Web Cars}\end{tabular} & \begin{tabular}[c]{@{}c@{}}\rotatebox{70}{Stanford Dogs}\end{tabular} & \begin{tabular}[c]{@{}c@{}}\rotatebox{70}{Fruit92}\end{tabular} & \begin{tabular}[c]{@{}c@{}}\rotatebox{70}{CUB200}\end{tabular} & \begin{tabular}[c]{@{}c@{}}\rotatebox{70}{Food101}\end{tabular} & \begin{tabular}[c]{@{}c@{}}\rotatebox{70}{Food172}\end{tabular} & \begin{tabular}[c]{@{}c@{}}\rotatebox{70}{Food200}\end{tabular} & \begin{tabular}[c]{@{}c@{}}\rotatebox{70}{Food500}\end{tabular} & \begin{tabular}[c]{@{}c@{}}\rotatebox{70}{FGVC Aircraft}\end{tabular} & \begin{tabular}[c]{@{}c@{}}\rotatebox{70}{Veg200}\end{tabular} & \rotatebox{70}{Avg.} \\\midrule

CLIP   & 62.30                                               & 89.10                                                 & 70.70                                                    & 65.70                                                   & 63.80                                              & 64.50                                                   & 41.90                                              & 54.70                                             & 85.90                                              & 30.20                                                   & 49.60                                          & 42.50                                               & 24.90                                                   & 35.90                                             & 55.84\\
CoOp & 75.45                                               & 93.16                                                 & 95.83                                                    & 77.72                                                   & 70.52                                              & 73.02                                                   & 72.39                                              & 70.91                                             & 86.40                                              & 60.60& 57.78& 47.40& 38.69                                                   & 61.21                                             & 70.08\\
CoCoOp & 72.06                                               & \textcolor{blue}{93.89}& 90.88                                                    & 71.58                                                   & 69.07                                              & 70.46                                                   & 63.70                                              & 64.04                                             & 87.38                                              & 44.60                                                   & 54.19& 45.00& 34.22                                                   & 50.36                                             & 65.11
\\
KgCoOp & 71.87                                               & 92.96                                                 & 92.08                                                    & 72.14                                                   & 70.21                                              & 67.56                                                   & 67.21                                              & 65.79                                             & 87.11                                               & 50.96  & 56.00  & 45.98                                                & 34.10                                                    & 53.53                                             & 66.25\\
TCP & 79.79                                               & 92.99                                                 & 97.45                                                    & 83.70                                                   & 73.00                                              & 74.45                                                   & 80.43                                              & 78.58                                             & 87.32                                              & 73.63                                                   &60.82                                              & 54.90                                                & 44.56                                                   & 77.2                                              & 75.63\\
ProText &61.86 &92.72 &72.42 &66.77 &61.61 &55.04 &47.76 &54.32 &85.35 &35.51 &46.49 &35.60 &29.01 &34.38&55.63 \\
ATPrompt &76.44 &93.32 &96.97 &79.05 &70.71 &73.89 &76.75 &74.55 &85.73 &65.73 &57.89 &48.90 &40.37 &67.40&71.98 \\
SCPT   & \textcolor{blue}{80.81}& 93.38                                                 & \textcolor{blue}{97.72}& \textcolor{blue}{86.21}& \textcolor{blue}{74.48}& \textcolor{blue}{75.58}& \textcolor{blue}{81.27}& \textcolor{blue}{80.48}& \textcolor{blue}{87.45}& \textcolor{blue}{75.34}& \textcolor{blue}{60.88}& \textcolor{blue}{55.00}& \textcolor{blue}{46.41}& \textcolor{blue}{78.72}& \textcolor{blue}{76.70}\\ \bottomrule 
\end{tabular}
\end{table}

\subsection{ Few-Shot Learning Evaluations}
To validate that the SCPT method offers enhanced fine-grained semantic distinction compared to existing CoOp-style prompt templates,  we conducted few-shot classification across all 14 fine-grained image datasets using K\--shot labeled source images. Evaluations were performed on a standard test domain within the same class space as the training set.

Table \ref{fewshot} reports the performance of SCPT on 14 fine-grained datasets under the 16-shot setting, achieving an average accuracy of 76.70\%.  Compared with the state-of-the-art method TCP, SCPT consistently outperforms it on all benchmarks except Oxford Pets. This dataset contains the fewest fine-grained categories, which constrains the expressiveness of the induced semantic topology and thus limits the potential gains from semantic-aware tuning. Overall, SCPT achieves an average improvement of more than 1\% across all datasets.

\subsection{Base-to-Novel Generalization}
\begin{table*}[htbp]
   \caption{Comparison with SOTA methods on base-to-novel generalization (\%) in 16-shots. HM: Harmonic mean. }
    \label{tab-sota}
    \centering
    
    \footnotesize
\setlength{\tabcolsep}{2.5pt} 
\renewcommand{\arraystretch}{0.6}
    \subfloat[Average results]{
        \begin{minipage}{0.32\textwidth}
        \centering
        
        \begin{tabular}{ccc|c}
        \hline 
        Method& Base & Novel & HM \\ \hline 
        CLIP  & 60.09 & 63.88 & 61.93 \\
        CoOp  & 76.26 & 55.12 & 63.99 \\
        CoCoOp  & 71.54 & 62.86 & 66.92 \\
        KgCoOp & 72.39 & 63.50 & 67.65 \\
        TCP & 77.62 & 64.54 & 70.48\\ 
         ProText &60.34&61.91&61.12\\
        ATPrompt&76.85&56.10&64.86 \\
        \hline 
        \rowcolor{gray!20}SCPT & \textcolor{blue}{78.72} & \textcolor{blue}{64.91} & \textcolor{blue}{71.15} \\ \hline 
        \end{tabular}
        \end{minipage}}
    \hfill
    \subfloat[Dog Breed]{
        \begin{minipage}{0.32\textwidth}
        \centering
        
        \begin{tabular}{ccc|c}
       \hline 
        Method& Base & Novel & HM \\ \hline 
        CLIP & 71.80 & 71.60 & 71.70 \\
        CoOp  & 87.71 & 58.89 & 70.47 \\
        CoCoOp  & 83.75 & 74.39 & 78.79 \\
        KgCoOp  & 79.32 & 63.09 & 70.28 \\
        TCP  & \textbf{87.73} & 74.88 & 80.80 \\
                ProText & 72.55 & 71.00 & 71.77 
\\
        ATPrompt& 87.08 & 67.16 & 75.83 
\\
         \hline 
        \rowcolor{gray!20}SCPT & 87.66& \textbf{75.30} & \textbf{81.01} \\\hline 
        \end{tabular}
        \end{minipage}}
    \hfill
    \subfloat[Oxford Flowers]{
        \begin{minipage}{0.32\textwidth}
        \centering
        
        \begin{tabular}{ccc|c}
        \hline
        Method& Base & Novel & HM \\ \hline
        CLIP
& 72.08 & \textbf{77.80} & 74.83 \\
        CoOp
& 97.63 & 69.55 & 81.23 \\
        CoCoOp
& 94.87 & 71.75 & 81.71 \\
        KgCoOp
& 95.00 & 74.73 & 83.65 \\
        TCP
& 97.73 & 75.57 & 85.23 \\
                        ProText 
& 74.36 & 78.44 & 76.35 
\\
        ATPrompt
& 98.02 & 69.28 & 81.18 
\\\hline
        \rowcolor{gray!20}SCPT & \textbf{98.08}& 75.89 & \textbf{85.57} \\\hline
        \end{tabular}
        \end{minipage}}
    \\[-1em]

    \subfloat[Oxford Pets]{
        \begin{minipage}{0.32\textwidth}
        \centering
        
        \begin{tabular}{ccc|c}
        \hline
        Method& Base & Novel & HM \\ \hline
        CLIP & 91.17 & 97.26 & 94.12 \\
        CoOp & 94.24 & 96.66 & 95.43 \\
        CoCoOp & \textbf{95.20} & 97.69 & \textbf{96.43} \\
        KgCoOp & 94.65 & 97.76 & 96.18 \\
         TCP & 94.67 & 97.20 & 95.92 \\
         ProText & 94.95 & 98.00 & 96.45 
\\
        ATPrompt & 95.61 & 97.04 & 96.32 
\\\hline
        \rowcolor{gray!20}SCPT & 94.76& \textbf{97.87} & 96.29 \\\hline
        \end{tabular}
        \end{minipage}}
    \hfill
    \subfloat[Stanford Cars]{
        \begin{minipage}{0.32\textwidth}
        \centering
        
        \begin{tabular}{ccc|c}
        \hline
        Method& Base & Novel & HM \\ \hline
        CLIP & 63.37 & 74.89 & 68.65 \\
        CoOp & 76.20 & 69.14 & 72.49 \\
        CoCoOp & 70.49 & 73.59 & 72.01 \\
        KgCoOp & 71.76 & 75.04 & 73.36 \\
        TCP & 80.80 & 74.13 & 77.32 \\
         ProText & 64.54 & 76.08 & 69.84 
\\
        ATPrompt & 77.27 & 66.54 & 71.50 
\\\hline
        \rowcolor{gray!20}SCPT & \textbf{81.17}& \textbf{75.66} & \textbf{78.32} \\\hline
        \end{tabular}
        \end{minipage}}
    \hfill
    \subfloat[Fruit92]{
        \begin{minipage}{0.32\textwidth}
        \centering
        
        \begin{tabular}{ccc|c}
        \hline
        Method& Base & Novel & HM \\ \hline
        CLIP & 53.00 & 61.10 & 56.76 \\
        CoOp & 85.39 & 46.96 & 60.60 \\
        CoCoOp & 74.58 & 60.05 & 66.53 \\
        KgCoOp & 76.73 & 57.37 & 65.65 \\
        TCP & 85.83 & 63.79 & 73.19 \\
         ProText & 54.63 & 52.70 & 53.65 
\\
        ATPrompt & 85.92 & 47.21 & 60.94 
\\\hline
        \rowcolor{gray!20}SCPT & \textbf{86.28}& \textbf{64.34} & \textbf{73.71} \\\hline
        \end{tabular}
        \end{minipage}}
    \\[-1em]

    \subfloat[CUB200]{
        \begin{minipage}{0.32\textwidth}
        \centering
        
        \begin{tabular}{ccc|c}
        \hline
        Method& Base & Novel & HM \\ \hline
        CLIP & 63.90 & 51.90 & 57.28 \\
        CoOp & 81.48 & 37.37 & 51.24 \\
        CoCoOp & 74.46 & 51.20 & 60.68 \\
        KgCoOp & 75.62 & 52.60 & 62.04 \\
        TCP & 82.10 & 52.97 & 64.39 \\
        ProText & 65.07 & 50.70 & 56.99 
\\
        ATPrompt & 82.02 & 40.96 & 54.64 
\\\hline
        \rowcolor{gray!20}SCPT & \textbf{82.96}& 52.67 & \textbf{64.43} \\\hline
        \end{tabular}
        \end{minipage}}
    \hfill
    \subfloat[Food101]{
        \begin{minipage}{0.32\textwidth}
        \centering
        
        \begin{tabular}{ccc|c}
        \hline
        Method& Base & Novel & HM \\ \hline
        CLIP & 90.10 & 91.22 & 90.66 \\
        CoOp & 89.44 & 87.50 & 88.46 \\
        CoCoOp & \textbf{90.70} & 91.70 & 91.09 \\
        KgCoOp & 90.50 & 91.70 & 91.09 \\
        TCP & 90.57 & 91.37 & 90.97 \\
        ProText & 90.20 & 91.98 & 91.08 
\\
        ATPrompt & 88.86 & 88.95 & 88.90 
\\\hline
        \rowcolor{gray!20}SCPT & 90.69& 91.54 & \textbf{91.15} \\\hline
        \end{tabular}
        \end{minipage}}
    \hfill
    \subfloat[Veg200]{
        \begin{minipage}{0.32\textwidth}
        \centering
        
        \begin{tabular}{ccc|c}
        \hline
        Method& Base & Novel & HM \\ \hline
        CLIP & 44.90 & 43.80 & 44.34 \\
        CoOp \cite{zhou2022learning}& 78.19 & 30.20 & 43.57 \\
        CoCoOp & 67.19 & 45.79 & 54.46 \\
        KgCoOp & 66.85 & 44.08 & 53.13 \\
        TCP & 83.27 & 49.53 & \textbf{62.11} \\
        ProText & 44.99 & 39.20 & 41.90 
\\
        ATPrompt & 79.85 & 32.53 & 46.23 
\\\hline
        \rowcolor{gray!20}SCPT & \textbf{84.23}& 48.59 & 61.63 \\\hline
        \end{tabular}
        \end{minipage}}
    \\[-1em]

    \subfloat[Web Cars]{
        \begin{minipage}{0.32\textwidth}
        \centering
        
        \begin{tabular}{ccc|c}
        \hline
        Method& Base & Novel & HM \\ \hline
        CLIP & 61.80 & 73.60 & 62.45 \\
        CoOp & 68.04 & 56.98 & 65.74 \\
        CoCoOp & 67.87 & 73.55 & 70.86 \\
        KgCoOp & 68.82 & 74.31 & 71.46 \\
        TCP & 68.70 & 72.50 & 70.55 \\
        ProText & 60.89 & 72.70 & 66.27 
\\
        ATPrompt & 67.69 & 62.27 & 64.87 
\\\hline
        \rowcolor{gray!20}SCPT & \textbf{69.85}& 73.16 & \textbf{71.47} \\\hline
        \end{tabular}
        \end{minipage}}
    \hfill
    \subfloat[Stanford Dogs]{
        \begin{minipage}{0.32\textwidth}
        \centering
        
        \begin{tabular}{ccc|c}
        \hline
        Method& Base & Novel & HM \\ \hline
        CLIP & 61.90 & 63.00 & 62.45 \\
        CoOp & 75.80 & 58.03 & 65.74 \\
        CoCoOp & 74.12 & 67.87 & 70.86 \\
        KgCoOp & 71.24 & 66.12 & 68.58 \\
        TCP & 75.45 & 68.79 & 71.97 \\
         ProText & 63.43 & 66.80 & 65.07 
\\
        ATPrompt & 76.5& 58.31& 66.18 
\\\hline
        \rowcolor{gray!20}SCPT & \textbf{75.89}& \textbf{69.67} & \textbf{72.65} \\\hline
        \end{tabular}
        \end{minipage}}
    \hfill
    \subfloat[FGVC Aircraft]{
        \begin{minipage}{0.32\textwidth}
        \centering
        
        \begin{tabular}{ccc|c}
        \hline
        Method& Base & Novel & HM \\ \hline
        CLIP & 27.19 & \textbf{36.29} & 31.09 \\
        CoOp & 39.24 & 30.49 & 34.30 \\
        CoCoOp & 33.41 & 23.71 & 27.74 \\
        KgCoOp & 36.21 & 33.55 & 34.83 \\
        TCP & 41.97 & 34.43 & 37.83 \\
        ProText & 30.91 & 34.13 & 32.44 
\\
        ATPrompt & 39.80& 30.74& 34.69 
\\\hline
        \rowcolor{gray!20}SCPT & \textbf{42.57}& 35.53 & \textbf{38.73} \\\hline
        \end{tabular}
        \end{minipage}}
    \\[-1em]

    \subfloat[Food200]{
        \begin{minipage}{0.32\textwidth}
        \centering
        
        \begin{tabular}{ccc|c}
        \hline
        Method& Base & Novel & HM \\ \hline
        CLIP & 55.20 & 59.00 & 57.03 \\
        CoOp & 64.96 & 52.31 & 57.95 \\
        CoCoOp & 62.17 & \textbf{59.87} & 61.00 \\
        KgCoOp & 64.22 & 59.17 & 61.59 \\
        TCP & 66.93 & 58.93 & 62.68 \\
        ProText & 52.25 & 55.80 & 53.97 
\\
        ATPrompt & 64.82 & 52.40 & 57.95 
\\\hline
        \rowcolor{gray!20}SCPT & \textbf{67.04}& 59.22 & \textbf{62.89} \\\hline
        \end{tabular}
        \end{minipage}}
    \hfill
    \subfloat[Food500]{
        \begin{minipage}{0.32\textwidth}
        \centering
        
        \begin{tabular}{ccc|c}
        \hline
        Method& Base & Novel & HM \\ \hline
        CLIP & 48.80 & \textbf{52.00} & 50.35 \\
        CoOp & 56.95 & 44.99 & 50.27 \\
        CoCoOp & 53.12 & 51.58 & 52.34 \\
        KgCoOp & 55.04 & 50.25 & 52.54 \\
        TCP & 61.37 & 49.62 & 54.87 \\
        ProText & 44.30 & 43.20 & 43.74 
\\
        ATPrompt & 57.85 & 43.91 & 49.93 
\\\hline
        \rowcolor{gray!20}SCPT & \textbf{61.72}& 49.48 & \textbf{54.93} \\\hline
        \end{tabular}
        \end{minipage}}
    \hfill
    \subfloat[Food172]{
        \begin{minipage}{0.32\textwidth}
        \centering
        
        \begin{tabular}{ccc|c}
        \hline
        Method& Base & Novel & HM \\ \hline
        CLIP & 36.00 & \textbf{40.80} & 38.25 \\
        CoOp & 72.28 & 32.63 & 44.96 \\
        CoCoOp & 49.76 & 40.04 & 44.37 \\
        KgCoOp & 63.22 & 39.06 & 48.29 \\
        TCP & 78.34 & 39.87 & 52.85 \\
                ProText & 31.72 & 36.00 & 33.72 
\\
        ATPrompt & 74.62 & 28.10 & 40.83 
\\
        \hline
        \rowcolor{gray!20}SCPT & \textbf{79.06}& 39.84 & \textbf{52.98} \\\hline
        \end{tabular}
        \end{minipage}}
\end{table*}

To ensure consistency with prior studies such as TCP,  we partition each dataset into two disjoint sets: base classes and novel classes,  following a zero-shot learning framework where base and novel classes do not overlap. This setup enables us to quantify the model's generalization capability to fine-grained,  out-of-domain categories. For each evaluation stage, SRE is computed from the textual semantic representations of the candidate class names and projected using the same random projection function as in training; the resulting SRE is then combined with the class name and the learnable context vectors to form the final textual prompt.

Table \ref{tab-sota} compares SCPT with CLIP and recent prompt tuning methods under the 16-shot setting. SCPT outperformed existing approaches,  showing improvements on both base and novel classes. The SRE-enhanced prompt captured semantic relation,  yielding a 1.10\% average gain over TCP across  14 datasets. The incorporation of Semantic Condensation loss  (ScLoss) mitigated irrelevant semantic interference,  resulting in a 71.15\% harmonic mean,  demonstrating strong generalization to novel classes.
In contrast to CoOp methods,  which often overfit to base classes,  SCPT achieved an 77.84\% improvement in harmonic mean performance,  a 9.79\% increase in novel class generalization,  and a 2.46\% boost in base class accuracy. These results highlight SCPT's superior ability to balance generalization and base class performance,  achieving the highest overall accuracy across all 14 datasets.

\begin{table}[!htbp]
\centering
\footnotesize
\setlength{\tabcolsep}{0.85pt}
\caption{ Effects of SRE and ScLoss combinations  (\%). SR:Semantic Relation Encoding. Sc:Semantic Condensation loss. TCP (Reg.) denotes the ablation reference baseline, i.e., CoOp-style prompt tuning with semantic regularization under the same experimental configuration.}
\label{tab:joint}
\renewcommand{\arraystretch}{1}

\begin{tabular}{lccccccccccccccc}
\hline
Method &\begin{tabular}[c]{@{}c@{}}\rotatebox{70}{Dog Breed}\end{tabular} & \begin{tabular}[c]{@{}c@{}}\rotatebox{70}{Oxford Pets}\end{tabular} & \begin{tabular}[c]{@{}c@{}}\rotatebox{70}{Oxford Flowers}\end{tabular} & \begin{tabular}[c]{@{}c@{}}\rotatebox{70}{Stanford Cars}\end{tabular} & \begin{tabular}[c]{@{}c@{}}\rotatebox{70}{Web Cars}\end{tabular} & \begin{tabular}[c]{@{}c@{}}\rotatebox{70}{Stanford Dogs}\end{tabular} & \begin{tabular}[c]{@{}c@{}}\rotatebox{70}{Fruit92}\end{tabular} & \begin{tabular}[c]{@{}c@{}}\rotatebox{70}{CUB200}\end{tabular} & \begin{tabular}[c]{@{}c@{}}\rotatebox{70}{Food101}\end{tabular} & \begin{tabular}[c]{@{}c@{}}\rotatebox{70}{Food172}\end{tabular} & \begin{tabular}[c]{@{}c@{}}\rotatebox{70}{Food200}\end{tabular} & \begin{tabular}[c]{@{}c@{}}\rotatebox{70}{Food500}\end{tabular} & \begin{tabular}[c]{@{}c@{}}\rotatebox{70}{FGVC Aircraft}\end{tabular} & \begin{tabular}[c]{@{}c@{}}\rotatebox{70}{Veg200}\end{tabular} & \rotatebox{70}{Avg.} \\
\hline
  TCP (Reg.)    &79.79    & 92.99 & 97.45   & 83.70        & 73.00  & 74.45        & 80.43   & 78.58  & 87.32&73.63& 60.82 &54.90    & 44.56    & 77.20  & 75.63            \\ 

SR       & 80.08    & 93.24 & 97.39   & 83.64        & 72.86  & 74.87        & 80.37   & 78.84  & 87.52&75.12&60.80&\textcolor{blue}{55.09}   & 43.99    & 77.38  & 75.80 \\
Sc & 79.91    & 93.33 & 97.64   & 84.04        & 73.30  & 74.80        & 80.50   & 79.65  & \textcolor{blue}{87.48}&74.42&\textcolor{blue}{60.92}&54.92   & 44.82    & 78.08  & 75.98   \\
SR \& Sc & \textcolor{blue}{80.81}& \textcolor{blue}{93.38}& \textcolor{blue}{97.72} &\textcolor{blue}{86.21} &\textcolor{blue}{74.48} &\textcolor{blue}{75.58} &\textcolor{blue}{81.27}& \textcolor{blue}{80.48} &87.45 &\textcolor{blue}{75.34}& 60.88 &55.00 &\textcolor{blue}{46.41} &\textcolor{blue}{78.72}&\textcolor{blue}{76.70}\\\bottomrule
\end{tabular}
\end{table}

\subsection{Ablation Study} \label{sec:ablation}
\noindent\textbf{Effects of SRE and ScLoss.}
The proposed SCPT method integrates two core components: SRE and ScLoss.
SRE encodes class names to reflect inter-class relationships and enhance class distinction,
while ScLoss suppresses irrelevant semantic information to improve robustness.
To ensure a fair and controlled evaluation of the proposed components, we adopt CoOp-style prompts with semantic regularization following the TCP formulation as the baseline setting in the ablation study. This design choice allows us to isolate the individual contributions of SRE and ScLoss under a consistent semantic regularization framework, avoiding performance variations caused by changes in the underlying prompt formulation.

As shown in Table~\ref{tab:joint}, SRE and ScLoss contribute to performance improvement through different mechanisms.
SRE enhances class separability by encoding fine-grained inter-class similarity structures, but its reliance on semantic relations makes it more vulnerable to semantic noise.
In contrast, ScLoss improves robustness by suppressing irrelevant variations via low-rank filtering, yet lacks the capacity to explicitly capture structural relations among classes.
When combined, the two components complement each other---SRE promotes discriminability, while ScLoss mitigates noise sensitivity---resulting in an average performance gain exceeding 1\%.
These results highlight their strong complementarity and validate the effectiveness of joint optimization.

\begin{figure}[htbp]
    \centering
    \includegraphics[width=0.67\linewidth]{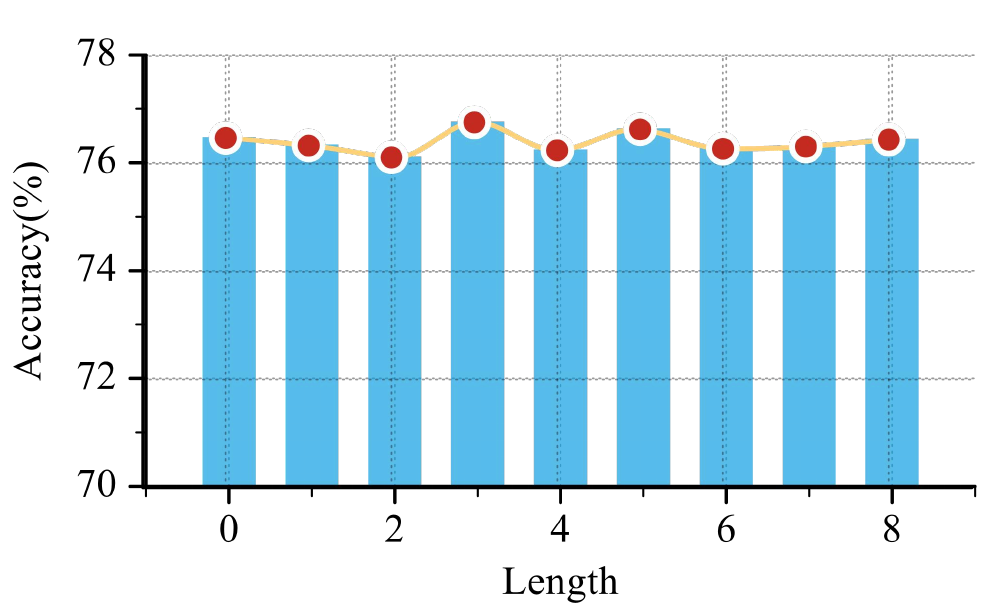}
    \caption{Impact of additional free space $d_{free}$. }
    \label{fig-df}

\end{figure}

\noindent\textbf{Effect of $d_{free}$ in SRE.} 
This study examines the impact of the additional free space $ d_{\text{free}} $ on encoding length and recognition accuracy in a 16-shot setting across 14 datasets,  as shown in Figure  \ref{fig-df}. We evaluate $ d_{\text{free}} $ values from 0 to 8,  with results indicating that $ d_{\text{free}} = 3 $ yields the best performance. However,  the accuracy difference between the highest  and lowest  values is only 0.36\%,  suggesting that recognition accuracy is not highly sensitive to variations in $ d_{\text{free}} $. This implies that the dimensionality reduction  is not strongly dependent on the reduced space size in the absence of conflicts.

\begin{figure}[!h]
    \centering
    \includegraphics[width=0.67\linewidth]{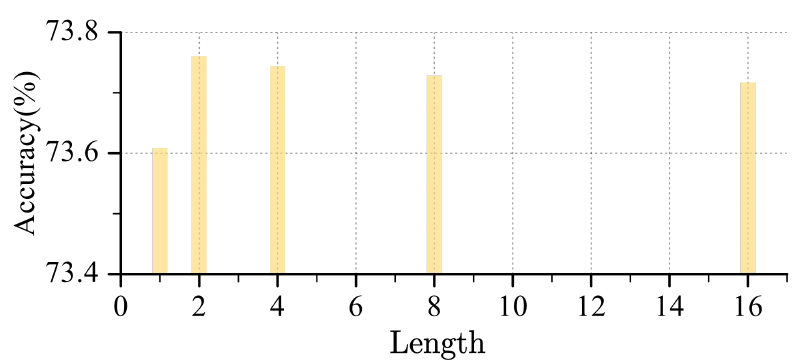}
    \caption{Effect of learnable prompt Length $M$.}
    \label{fig-lm}
     \vspace{-1em}
\end{figure}
\noindent\textbf{Effect of Prompt length M.}
In a 4-shot experiment,  we investigate the impact of prompt length $ M $ on few-shot performance,  as shown in Figure~\ref{fig-lm}. We evaluate prompt lengths of 1,  2,  4,  8,  and 16,  finding that SRE-encoded prompts are largely insensitive to learnable parameter variations. While a prompt length of 2 yields the highest accuracy,  the difference from the lowest result is just 0.16\%. Shorter prompt lengths generally outperform longer ones,  indicating that SRE encoding inherently captures more class-specific information,  which is particularly advantageous in fine-grained recognition tasks.

\begin{figure}[htbp]
\centering
\includegraphics[width=0.568\linewidth]{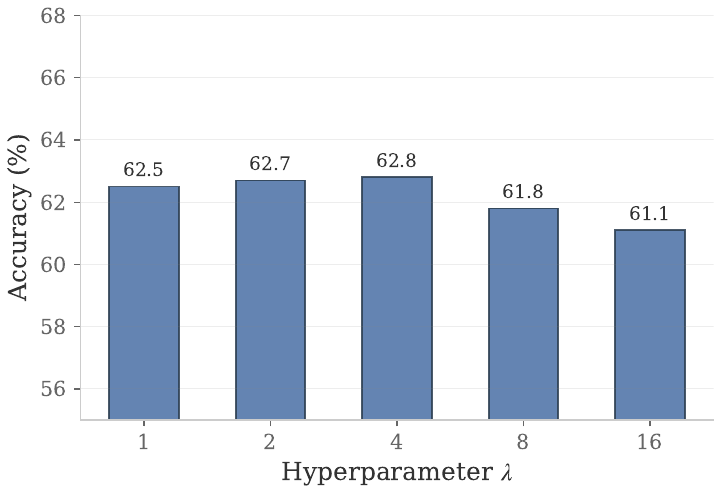}
\caption{Effect of loss weight $\lambda$.}
\label{fig:lambda}
\vspace{-1.5em}
\end{figure}

\noindent\textbf{Effect of Hyperparameter $\lambda$.}
 We evaluate the sensitivity of the loss weight $\lambda$ by measuring average accuracy on four fine-grained datasets, namely Fruit92, FGVC-Aircraft, CUB200, and Stanford Dogs, using the ViT-B/32 backbone.
The value of $\lambda$ is varied over $\{1, 2, 4, 8, 16\}$.
As shown in Figure~\ref{fig:lambda}, the performance remains stable when $\lambda$ ranges from 1 to 8, with accuracy variations below 1\%.
The highest accuracy is achieved at $\lambda=4$, while $\lambda=8$ yields a comparable result.
For consistency with prior work and to avoid dataset-specific tuning, we use $\lambda=8$ in all experiments.

\begin{figure}[htbp]
\centering
\includegraphics[width=0.568\linewidth]{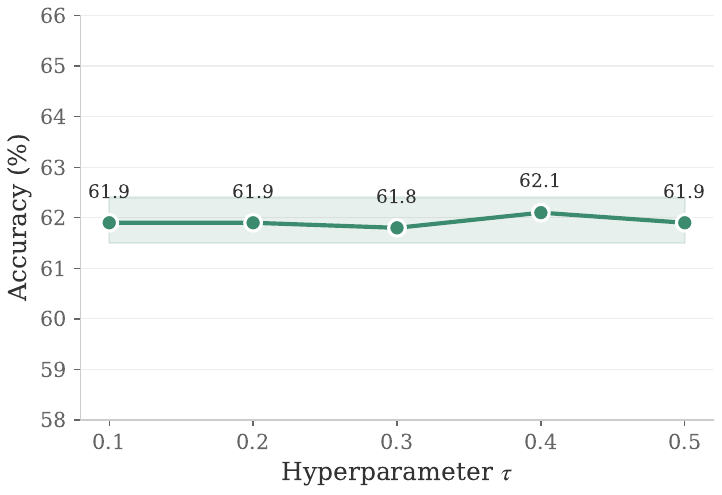}
\caption{Effect of threshold $\tau$.}
\label{fig:tau}
\vspace{-1.5em}
\end{figure}

\noindent\textbf{Effect of Threshold $\tau$.}
The threshold $\tau$ controls the strength of feature noise filtering.
We evaluate the sensitivity to $\tau$ by varying its value from 0.1 to 0.5 under the same experimental setting, using the ViT-B/32 backbone on Fruit92, FGVC-Aircraft, CUB200, and Stanford Dogs.
As shown in Figure~\ref{fig:tau}, the average classification accuracy remains nearly unchanged across the entire range, with variations below 0.3\%.
Based on this observation, we use a fixed value of $\tau$ in all experiments.

\noindent\textbf{Sensitivity to Random Projection Initialization.}
 We evaluate the impact of initialization on the projection matrix $\mathbf{W}$ using five random seeds on the Fruit92 dataset (ViT-B/32). As shown in Table \ref{random}, SCPT exhibits remarkable robustness, yielding a mean accuracy of 75.3\% $\pm 0.23$. With a variation of less than 0.6\% across all trials, the results conclusively demonstrate that the proposed signed random projection strategy is insensitive to initialization noise, validating its reliability for consistent deployment.

\begin{table}[h]
    \centering
    \caption{Sensitivity analysis of the random projection matrix $\mathbf{W}$ initialization on the Fruit92 dataset with a ViT-B/32 backbone. }
   
    \begin{tabular}{lccccccc}
        \toprule
        \textbf{Random Seed} & 1 & 2 & 3 & 42 & 100 & \textbf{Mean} & \textbf{Std} \\
        \midrule
        \textbf{Accuracy (\%)}   & 75.4 & 75.3 & 75.4 & 75.5 & 74.9 & 75.3 & $\pm$0.23 \\
        \bottomrule
    \end{tabular}
    \label{random}
\end{table}

\noindent\textbf{Comparison of Computational Efficiency.} 
  Table \ref{tab:efficiency} summarizes the complexity analysis on the large-scale Food-500 dataset. Results indicate that while SCPT incorporates the SRE module, it maintains a lightweight design with 0.79M learnable parameters, representing merely 0.77\% of the total model capacity. In terms of computational speed, SCPT achieves a training throughput of 140.1 img/s, which is more than 20$\times$ faster than CoCoOp and comparable to both CoOp and TCP. During inference, SCPT achieves 228.4 ms per batch, remaining close to TCP in latency while maintaining a similar memory footprint. These results suggest that SCPT maintains a similar parameter scale and computational efficiency to TCP, while providing consistent performance improvements.
\begin{table}[htbp]
\centering
\caption{Computational efficiency and model complexity analysis on Food-500.}
\label{tab:efficiency}
\resizebox{\linewidth}{!}{%
\begin{tabular}{lcccccc}
\toprule
\multirow{2}{*}{\textbf{Method}} & \multicolumn{2}{c}{\textbf{Model Complexity}} & \multicolumn{4}{c}{\textbf{Computational Efficiency}} \\
\cmidrule(lr){2-3} \cmidrule(lr){4-7}
 & \textbf{Params (M)} & \textbf{Ratio (\%)} & \textbf{Init. (s)} & \textbf{Train (img/s)} $\uparrow$ & \textbf{Infer. (ms)} $\downarrow$ & \textbf{Mem. (MB)} \\
\midrule
CoOp \cite{zhou2022learning} & 0.004 & 0.004 & 0.058 & 174.9& 183.0 & 941.4 \\
CoCoOp \cite{zhou2022conditional} & 0.103 & 0.100 & 0.063 & 6.2 & 5135.5 & 942.8 \\
TCP \cite{yao2024tcp} & 0.793 & 0.780 & 0.127 & 170.8& 264.1 & 955.6 \\
\textbf{SCPT (Ours)} & 0.790 & 0.770 & 0.286 & 140.1 & 228.4 & 947.1 \\
\bottomrule
\end{tabular}%
}
\end{table}

\begin{table}[!h]
  \centering
  \caption{Performance Comparison with Different Backbones (ViT-B/32 and RN50)}
  \label{tab:performance_comparison}
  \setlength{\tabcolsep}{6pt} 
  \renewcommand{\arraystretch}{0.7} 
  \begin{tabular}{lcccccc}
    \toprule
    Method & Fruit92 & Veg200 & Webcar & Aircraft & Dog Breed & Avg. \\
    \midrule
    \multicolumn{7}{l}{\textit{\textbf{Backbone: ViT-B/32}}} \\ 
    \midrule
    CLIP     & 48.95 & 37.42 & 59.43 & 20.85 & 59.35 & 45.20 \\
    CoOp     & 70.14 & 58.49 & 62.90 & 29.93 & 70.23 & 58.34 \\
    CoCoOp   & 30.47 & 28.90 & 48.43 & 7.00  & 55.07 & 33.97 \\
    KgCoOp   & 63.22 & 51.31 & 63.84 & 26.70 & 67.13 & 54.44 \\
    TCP      & 69.60 & 64.13 & 64.00 & 28.37 & 68.67 & 58.95 \\
    ProText  & 42.60 & 32.13 & 57.52 & 10.11 & 57.17 & 39.91 \\
    ATPrompt & 71.40 & 61.17 & 62.80 & 30.60 & 71.30 & 59.45 \\
    SCPT     & \textbf{74.40} & \textbf{70.74} & \textbf{66.97} & \textbf{31.93} & \textbf{73.13} & \textbf{63.43} \\
    \midrule
    \multicolumn{7}{l}{\textit{\textbf{Backbone: RN50}}} \\ 
    \midrule
    CLIP     & 46.32 & 35.14 & 54.66 & 18.72 & 55.90 & 42.15 \\
    CoOp     & 67.63 & 54.76 & 59.43 & 28.35 & 67.01 & 55.44 \\
    CoCoOp   & 37.80 & 28.40 & 49.07 & 16.27 & 50.83 & 36.47 \\
    KgCoOp   & 60.63 & 47.90 & 59.93 & 26.05 & 63.63 & 51.63 \\
    TCP      & 67.63 & 63.88 & 60.94 & 27.54 & 66.16 & 57.23 \\
    ProText  & 38.45 & 28.70 & 52.05 & 6.00  & 49.08 & 34.86 \\
    ATPrompt & 69.47 & 58.07 & 59.07 & 28.33 & 67.27 & 56.44 \\
    SCPT     & \textbf{73.57} & \textbf{68.53} & \textbf{62.99} & \textbf{30.62} & \textbf{70.63} & \textbf{61.27} \\
    \bottomrule
  \end{tabular}
\end{table}

\noindent\textbf{ Robustness Across Backbones.}   While our main experiments are conducted with ViT-B/16, we further evaluate SCPT on two additional visual backbones, ViT-B/32 and ResNet-50, to assess its generality across architectures. Experiments are performed on five representative fine-grained datasets: Fruit92, Veg200, WebCar, FGVC Aircraft, and Dog Breed. As shown in Table~\ref{tab:performance_comparison}. SCPT achieves consistent state-of-the-art performance on both ViT-B/32 and ResNet-50, exceeding the strongest prior methods by a clear margin in average accuracy. This confirms that the effectiveness of SCPT generalizes well across both transformer- and CNN-based visual backbones.

\subsection{Comparison with Visual Prompt Tuning Methods}

\begin{table}[h]
\centering
\caption{Comparison of prompt learning methods with respect to prompt type, depth, modality, and average accuracy across 15 datasets. 
\textit{Depth} denotes the number of transformer layers where learnable prompts are injected.}
\label{tab:prompt_comparison}
\begin{tabular}{lcccc}
\toprule
Method & Prompt Type & Depth & Modalities & Avg Acc (\%) \\
\midrule
MaPLe & Deep & 9 & Text + Vision & 62.79 \\
PromptSRC & Deep & 9 & Text + Vision & 76.94 \\
\textbf{SCPT (ours)} & Shallow & 1 & Text-only & \textbf{76.70} \\
\bottomrule
\end{tabular}
\end{table}
  Although SCPT is designed as a purely textual prompt tuning method, we include a comparison with representative recent multi-modal prompt tuning approaches to analyze the relationship between prompt complexity and recognition performance.
As shown in Table~\ref{tab:prompt_comparison}, MaPLe and PromptSRC adopt deep prompt injection across nine transformer layers and leverage both visual and textual modalities, whereas SCPT employs a single-layer shallow prompt and operates exclusively in the textual domain.

Despite this substantially lighter design, SCPT attains an average accuracy of 76.70\%, which is comparable to PromptSRC in terms of average accuracy and markedly higher than MaPLe across 14 datasets.
These results indicate that strong performance can be achieved with shallow, text-only prompt tuning, suggesting that explicitly modeling semantic structure in textual prompts is a viable alternative to deep or multi-modal prompt designs.

\subsection{Visualization}
\begin{figure*}[h]
\centering
\includegraphics[width=\linewidth]{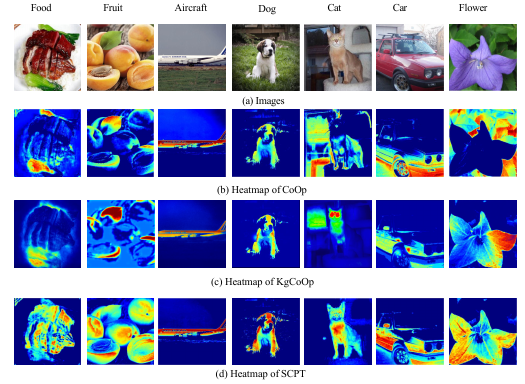}
\caption{Comparison of visual attention heatmaps generated by different methods:  (a) Ground-truth images;  (b) Heatmaps from CoOp similarity scores;  (c) Heatmaps from KgCoOp similarity scores; (d) Heatmaps from SCPT similarity scores.}
\label{fig-visual2}
\vspace{-1.5em}
\end{figure*}

  To investigate the impact of textual prompts on image feature representations, 
we compute similarity scores between target images and textual prompts using 
CoOp, KgCoOp, and the proposed SCPT method. Grad-CAM \cite{selvaraju2017grad} is then applied to backpropagate these similarity scores 
through the image encoder, producing gradient-based heatmaps that visualize image regions 
most sensitive to the target similarity scores, as shown in Figure~\ref{fig-visual2}.

Textual embeddings act as explicit supervisory signals during backpropagation, 
guiding gradient responses toward the target class and facilitating more discriminative 
region localization.   Compared with CoOp and KgCoOp, SCPT produces heatmaps that exhibit 
stronger focus on target regions, clearer foreground-background separation, and finer 
class-specific details, while suppressing irrelevant visual information.

\section{Limitations}

While our proposed framework demonstrates strong effectiveness in text-based prompt learning, several limitations remain. 
First, our method focuses exclusively on textual prompts and does not incorporate learnable visual prompts. 
Recent works such as MaPLe~\cite{khattak2023maple} and PromptSRC~\cite{khattak2023self} have shown that jointly optimizing textual and visual prompts can further enhance adaptation and generalization in certain scenarios. 
Extending SCPT to a multimodal prompt learning framework may provide additional representational flexibility, which we leave for future work.

Second, our current design adopts shallow prompt tuning and injects learnable prompts only at a limited number of layers. 
Although this design choice helps maintain parameter efficiency and training stability, it does not explore deeper prompt mechanisms that interact with intermediate layers of the backbone, as investigated in VPT~\cite{jia2022visual} and DualPrompt~\cite{wang2022dualprompt}. 
Incorporating deep prompts into the proposed structured and condensed prompt formulation is a promising direction for future research. 

Finally, Although SCPT is formulated as a complete prompt tuning model within the CoOp-style framework, extending it to other prompt learning paradigms is an interesting direction for future work.

\section{Conclusion}
This work revisits prompt tuning for fine-grained recognition from a structural perspective and argues that treating category prompts as isolated tokens fundamentally limits the ability of vision--language models to discriminate subtle inter-class differences. To address this issue, we propose Structured-Condensed Prompt Tuning (SCPT), which explicitly incorporates global semantic structure into textual prompt learning. By introducing Semantic Relation Encoding (SRE), SCPT captures inter-class semantic topology in a lightweight and parameter-efficient manner, while the Semantic Condensation loss (ScLoss) further refines supervision by suppressing redundant or noisy semantic components in the shared embedding space.

Extensive experiments on 14 fine-grained benchmarks demonstrate that SCPT consistently improves both few-shot adaptation and base-to-novel generalization, indicating that structured semantic modeling is particularly beneficial under limited data and distribution shift. The results suggest that incorporating global semantic relations provides a more effective inductive bias than treating category prompts as independent tokens.

Despite these advantages, SCPT has certain limitations. Its effectiveness depends on the availability of a sufficiently rich category set, as semantic topology becomes less informative when the number of classes is small. In addition, the current framework focuses exclusively on textual prompt learning and does not exploit complementary visual or multimodal prompting mechanisms. Future work will explore these directions and extend SCPT to large-scale and open-vocabulary recognition settings.

\appendix
\section{T-SNE visualizations of visual embeddings}\label{t-sne}
\begin{figure}[htbp]
    \centering
    \includegraphics[width=0.65\linewidth]{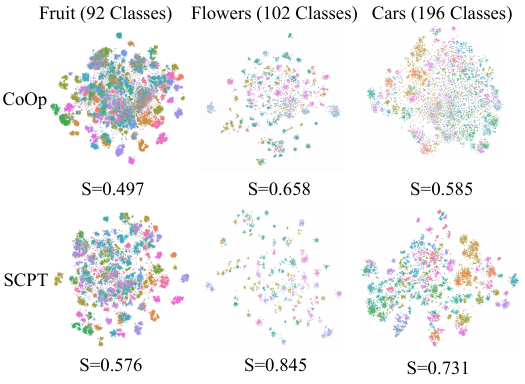}
    \caption{t-SNE visualizations of visual embeddings on three datasets. 
    The \textbf{Silhouette Coefficient (S)} is reported to quantify cluster compactness and separation.}
    \label{fig-visual}
     \vspace{-1.5em}
\end{figure}

To investigate the impact of SCPT on feature space distribution, we conduct a comparative analysis with the baseline method CoOp using t-SNE visualizations of visual embeddings extracted from test set samples across three datasets with varying numbers of categories: Fruit92 (92 classes), Oxford Flowers (102 classes), and Stanford Cars (196 classes), as shown in Figure~\ref{fig-visual}.

  The Silhouette Coefficient ($S$) is employed to quantitatively characterize the clustering structure of the visual embeddings, measuring the relative compactness of intra-class samples with respect to inter-class separation. Higher $S$ values indicate more coherent and well-separated clusters.

  As illustrated in Figure~\ref{fig-visual}, the baseline CoOp exhibits relatively diffuse cluster structures with indistinct boundaries across all evaluated datasets, which is reflected by its lower Silhouette Coefficients. In contrast, SCPT consistently produces more compact intra-class distributions and clearer separation between categories, yielding higher $S$ values. Furthermore, the performance gap between SCPT and CoOp becomes more pronounced as the number of categories increases, indicating that SCPT is effective in maintaining structured feature representations under large-scale and fine-grained recognition settings.

\section{Comparison with Conventional Few-Shot Learning Settings}

Conventional few-shot learning (FSL) methods are typically formulated under the $C$-way-$K$-shot episodic paradigm, where models are trained on base classes and evaluated on disjoint novel classes via repeated few-class classification tasks.
While widely adopted, this protocol imposes restrictive assumptions on task scale and data availability that diverge from real-world recognition settings.

Specifically, episodic FSL evaluates performance on tasks with a small number of classes (e.g., 5-way or 10-way), whereas practical applications often require discriminating among hundreds of categories simultaneously.
Moreover, episodic sampling emphasizes local class subsets and assumes abundant query samples, limiting its ability to reflect global inter-class confusion and large-scale recognition behavior.
Most critically, its closed-vocabulary evaluation assesses performance only within individual tasks, offering limited insight into open-vocabulary generalization.

To better align with large-scale and open-world scenarios, we adopt two complementary evaluation protocols commonly used in prompt-based and representation-learning methods: standard few-shot classification and base-to-novel generalization.
In the few-shot setting, the model is trained with $k$ labeled samples per class under a global supervised setup and evaluated on a held-out test set, directly measuring scalability under limited supervision.
In the base-to-novel setting, training is restricted to base classes, followed by evaluation on both base and novel categories, with generalization quantified by the harmonic mean of their accuracies.

\section{Theoretical Analysis of Signed Random Projection}

We provide a theoretical justification for the sign-based binarization by showing that the Hamming distance between binary projections consistently approximates the angular distance between the original semantic embeddings. Since vision--language models such as CLIP rely on cosine similarity, this result confirms that our binarization preserves the essential semantic topology of the embedding space.

\subsection{Equivalence between Hamming Distance and Angular Distance}

\textbf{Lemma 1 (Angular Distance Preservation).}
Let $S_i, S_j \in \mathbb{R}^N$ be two non-zero embedding vectors, and let
$W \in \mathbb{R}^{d \times N}$ be a random projection matrix whose rows are independently sampled from a spherically symmetric distribution.
Define the binary codes
$P_i = \operatorname{sign}(W S_i)$ and
$P_j = \operatorname{sign}(W S_j)$.
The normalized Hamming distance between the two codes is defined as:
\begin{equation}
d_H(P_i, P_j)
= \frac{1}{d} \sum_{k=1}^d \mathbf{1}[P_{ik} \neq P_{jk}],
\end{equation}
where $\mathbf{1}[\cdot]$ is the indicator function. This is an unbiased estimator of the normalized angular distance between $S_i$ and $S_j$, namely:
\begin{equation}
\mathbb{E}[d_H(P_i, P_j)]
= \frac{\theta_{ij}}{\pi},
\end{equation}
where $\theta_{ij} \in [0,\pi]$ denotes the angle between $S_i$ and $S_j$.

\textbf{Proof.}
Consider a single random projection vector $w$, corresponding to one row of $W$.
The induced hyperplane $\{x : w^\top x = 0\}$ separates $S_i$ and $S_j$ if and only if
$\operatorname{sign}(w^\top S_i) \neq \operatorname{sign}(w^\top S_j)$.
Due to the rotational invariance of the distribution of $w$, the probability of this event depends solely on the angle $\theta_{ij}$ between the two vectors.
It is a classical result in Locality Sensitive Hashing (LSH)  that a random hyperplane separates two vectors with probability $\theta_{ij}/\pi$.
Averaging over $d$ independent projections yields
$\mathbb{E}[d_H(P_i, P_j)] = \theta_{ij}/\pi$.

\subsection{Concentration of the Hamming Distance}

\textbf{Theorem 1 (Concentration).}
For any $\varepsilon > 0$, the empirical Hamming distance concentrates around its expectation as
\begin{equation}
\operatorname{Pr}\!\left(
\left| d_H(P_i, P_j) - \frac{\theta_{ij}}{\pi} \right|
\ge \varepsilon
\right)
\le
2 \exp(-2 d \varepsilon^2).
\end{equation}

\textbf{Proof.}
Each bit disagreement in the Hamming distance is an independent Bernoulli trial with success probability $\theta_{ij}/\pi$.
The normalized Hamming distance is therefore the empirical mean of $d$ i.i.d.\ Bernoulli variables.
Applying Hoeffding's inequality directly yields the stated bound.

\subsection{Rationale for Dimension Selection}

The projection dimension $d$ governs both the stability and the capacity of the Hamming space.
While signed random projection is inherently probabilistic, increasing $d$ reduces the variance of the angular distance estimation (as per Theorem 1) and lowers the probability of code collisions.
In practice, for $N$ semantic categories, we choose
\begin{equation}
d = \left\lceil \log_2 N + d_{\mathrm{free}} \right\rceil,
\end{equation}
where $d_{\mathrm{free}}$ is a small safety margin.
As shown in Figure~\ref{fig-df}, performance saturates once $d$ is sufficiently large to produce stable Hamming representations, indicating diminishing returns from further increasing the dimensionality.
\bibliographystyle{unsrtnat}
\bibliography{scpt}

\end{document}